\documentclass{article}

\usepackage[preprint,nonatbib]{neurips_2025}
\usepackage{amsmath}

\usepackage[utf8]{inputenc} % allow utf-8 input
\usepackage[T1]{fontenc}    % use 8-bit T1 fonts
\usepackage{hyperref}       % hyperlinks
\usepackage{verbatim}

\usepackage{url}            % simple URL typesetting
\usepackage{booktabs}       % professional-quality tables
\usepackage{amsfonts}       % blackboard math symbols
\usepackage{nicefrac}       % compact symbols for 1/2, etc.
\usepackage{microtype}      % microtypography
\usepackage{xcolor}         % colors
\usepackage{graphicx}

\title{ Mixture-of-Experts for Personalized and Semantic-Aware Next Location Prediction}
\author{Shuai Liu, Ning Cao, Yile Chen, Yue Jiang, Gao Cong\\
College of Computing and Data Science, Nanyang Technological University\\
50 Nanyang Avenue, Singapore, 639798 \\
}

\begin{document}

\maketitle

\vspace{-15pt}

\begin{abstract}
Next location prediction plays a critical role in understanding human mobility patterns. However, existing approaches face two core limitations: (1) they fall short in capturing the complex, multi-functional semantics of real-world locations; and (2) they lack the capacity to model heterogeneous behavioral dynamics across diverse user groups.
To tackle these challenges, we introduce NextLocMoE, a novel framework built upon large language models (LLMs) and structured around a dual-level Mixture-of-Experts (MoE) design.
Our architecture comprises two specialized modules: a Location Semantics MoE that operates at the embedding level to encode rich functional semantics of locations, and a Personalized MoE embedded within the Transformer backbone to dynamically adapt to individual user mobility patterns.
In addition, we incorporate a history-aware routing mechanism that leverages long-term trajectory data to enhance expert selection and ensure prediction stability.
Empirical evaluations across several real-world urban datasets show that NextLocMoE achieves superior performance in terms of predictive accuracy, cross-domain generalization, and interpretability.
\begin{comment}
Next location prediction is a key task in human mobility modeling. 
Existing methods face two challenges: (1) they fail to capture the multi-faceted semantics of real-world locations; and (2) they struggle to model diverse behavioral patterns across user groups.
To address these issues, we propose NextLocMoE, a large language model (LLM)-based framework for next location prediction, which integrates a dual-level Mixture-of-Experts (MoE) architecture. 
It comprises two complementary modules: a Location Semantics MoE at the embedding level to model multi-functional location semantics, and a Personalized MoE within LLM's Transformer layers  to adaptively capture user behavior patterns. 
To enhance routing stability and reliability, we introduce a historical-aware router that integrates long-term historical trajectories into expert selection.
Experiments on multiple real-world datasets demonstrate that NextLocMoE significantly outperforms existing methods in terms of accuracy, transferability, and interpretability.
\end{comment}
\end{abstract}

\vspace{-15pt}
% !TEX root = neurips_2025.tex
\section{Introduction}
With the advancement of mobile devices and  urban digitalization, predicting a user's next location based on his/her past trajectory has become a critical task across various domains, including intelligent transportation~\cite{liu2020real}, personalized service~\cite{li2024mcn4rec}, and urban management~\cite{yang2024cdrp3}.
The goal is to model user's mobility patterns and moving intentions for inferring his/her most likely next destination. 
%Accurate next location prediction is essential for intelligent service recommendation, for effective crowd flow analysis and urban-scale resource allocation.

%Researchers have developed various models for this task. 
Various methods have been proposed for this task. 
Early approaches relied on recurrent neural networks~\cite{chung2014empirical, graves2012long} to capture temporal dependencies.
With the emergence of Transformer~\cite{vaswani2017attention}, methods like MHSA~\cite{hong2023context}, CLLP~\cite{zhou2024cllp}, and GETNext~\cite{yang2022getnext}  were developed to capture complex spatiotemporal interactions.
Recently, large language models have been introduced into next location prediction.
Methods like Llama-Mob~\cite{tang2024instruction}, LLMMob~\cite{wang2023would}, and NextLocLLM~\cite{liu2024nextlocllm}  leverage LLMs' language understanding, reasoning ability, and pre-trained world knowledge to enhance predictive performance.
%seek to transfer LLMs' strong capabilities in language understanding and reasoning, exploiting their pre-trained world knowledge to enhance the effectiveness of next location prediction.

While existing methods have made notable progress, they still face two major challenges.
First, most models learn a single embedding for each location, which fails to capture the complex semantics that real-world locations often exhibit. 
For example, locations in city center may simultaneously serve commercial, residential, and educational purposes. 
Representing such multi-functional semantics with a single embedding limits the model’s expressiveness and predictive accuracy.
Second, most methods adopt a shared set of parameters for all users, which limits their ability to model behavioral heterogeneity across user groups. 
In real-world scenarios, distinct user groups—such as students, office workers, or tourists—exhibit different   spatial coverage and temporal routines.
While some models try to introduce personalized modeling via user embeddings~\cite{zhou2024cllp,yang2022getnext}, they face several limitations.
One limitation is their reliance on user IDs, which makes them vulnerable to cold-start problems when dealing with previously unseen users.
Another limitation is that 
user embeddings lack interpretability, offering little insight into behavioral patterns the model has learned.
%For instance, a local student and  a visitor may differ significantly in travel times, activity patterns, and destination preferences.
%The lack of personalized modeling hampers both accuracy and generalization. 
To tackle these challenges, we propose NextLocMoE, a dual-level Mixture-of-Experts (MoE) based LLM framework for next location prediction, which jointly models location semantics and user behavioral patterns.

To model location semantics, we design Location Semantics MoE, which enriches location representations by combining a shared spatial embedding with  expert embeddings specialized for different functional roles.
The shared embedding encodes geographic coordinates to capture general spatial features.
The routing mechanism of this MoE module activates the top-$k$ most relevant location function-specific experts, each encoding  the same coordinates into a function-aware embedding.
This results in multiple expert embeddings that reflect the diverse semantics a single location may exhibit. 
To inject semantic priors and improve interpretability, each expert is initialized with LLM-encoded natural language descriptions of predefined location function categories.

To capture user behavioral patterns, NextLocMoE integrates  Personalized MoE into selected Transformer layers of the LLM backbone by replacing the standard feedforward networks (FFNs). 
This design  enables group-level personalization while preserving LLM's  semantic encoding capacity. 
We predefine a set of user groups and encode their natural language descriptions using LLM to obtain  group-specific embeddings.
The router of Personalized MoE combines user’s historical trajectory representation with these embeddings to dynamically select the most relevant expert submodules.
Unlike the top-$k$ routing strategy used in Location Semantics MoE, Personalized MoE employs a confidence threshold based expert activation mechanism inspired by \cite{huang2024harder}.
This design is motivated by two considerations: (1) users may exhibit varying degrees of behavioral ambiguity, making it preferable to flexibly adjust the number of active experts; and (2) limiting expert activation reduces computational overhead. 
As a result, Personalized MoE activates only a few experts for users with consistent behavioral patterns, while allocating more capacity to users with uncertain or mixed behaviors.

%To enhance router’s global understanding of user behaviors and improve the stability of expert selection,  NextLocMoE introduces a historical-aware router that explicitly incorporates long-term historical trajectories during expert routing.
To improve long-term behavior awareness and expert selection stability in both MoE modules, NextLocMoE introduces a historical-aware router that explicitly incorporates historical trajectories into expert routing.
 In conclusion, our main contributions are summarized as follows:
 \begin{itemize}
     \item We propose NextLocMoE, the first LLM-based framework to integrate Mixture-of-Experts (MoE)  for next location prediction.
     It comprises (i) a Location Semantics MoE for modeling the multi-functional roles of  locations, and (ii) a Personalized MoE to capture user behavioral patterns. 
     Each module is  guided by expert-specific priors and customized routing strategy.
     \item We introduce a historical-aware router that incorporates long-term historical trajectory into expert selection, enhancing the contextual stability and reliability of  expert routing.
     \item Extensive experiments on multiple real-world datasets demonstrate that NextLocMoE consistently outperforms other baselines under both fully-supervised and zero-shot settings. 
     Case studies further highlight the model’s ability to provide interpretable predictions.

 \end{itemize}
% !TEX root = neurips_2025.tex
\section{Related Work}
\subsection{Next Location Prediction}
Next location prediction aims to forecast the most probable location a user will visit in the near future, based on his/her past trajectories. 
Early methods relied on recurrent neural networks like GRU~\cite{chung2014empirical} and LSTM~\cite{graves2012long}, to capture temporal dependencies. 
DeepMove~\cite{feng2018deepmove} enhances trajectory representation by jointly modeling short-term interests and long-term preferences.
However, these methods struggle with long-range dependencies and suffer from limited parallelism, which constrains their scalability.
With the rise of Transformer~\cite{vaswani2017attention}, attention-based methods have become the mainstream in next location prediction.
MHSA~\cite{hong2023context} models transition relationships between locations via multi-head self-attention. 
CLLP~\cite{zhou2024cllp} integrates local and global spatiotemporal contexts to better capture dynamic user interests.
GETNext~\cite{yang2022getnext} and SEAGET~\cite{al2025seaget} introduce global trajectory flow graphs and graph-enhanced Transformer models, leveraging collaborative mobility signals to improve predictive performance.
%These methods demonstrate substantial improvements in trajectory modeling over traditional RNN-based approaches.

In recent years, breakthroughs in large language models~\cite{achiam2023gpt,liu2024deepseek,touvron2023llama} have inspired researchers to explore their potential in next location prediction.
Llama-Mob~\cite{tang2024instruction} and LLMMob~\cite{wang2023would} design task-specific prompts, while NextLocLLM~\cite{liu2024nextlocllm} leverages  LLM as both a semantic enhancer and a predictor.
These methods exploit pre-trained world knowledge and reasoning capabilities to improve the semantic understanding of user mobility and enhance both prediction accuracy and generalization.

Despite these advances, two key limitations remain.
First, most models assign a single embedding to each location, which fails to capture the multifaceted semantics of real-world locations.
Second, most models use a shared set of parameters for all users, overlooking behavioral differences among user groups.
These limitations constrain both the accuracy and the adaptability of existing methods in real-world settings.
Therefore, we propose NextLocMoE, a novel framework that introduces a dual-level Mixture-of-Experts architecture to model both location semantics and  user behaviors.
For a more detailed discussion of related work on next location prediction, please refer to App.~\ref{sec:related app next loc}.

\subsection{Mixture of Experts}
Mixture of Experts (MoE) is  designed to enhance model expressiveness and computational efficiency.
It maintains a pool of expert networks and dynamically activates a subset of them for each input, allowing MoE-based models to achieve performance of larger architectures while keeping computation cost  low. 
MoE has achieved notable success in natural language processing, with prominent examples like GShard~\cite{lepikhin2020gshard}, Switch Transformer~\cite{fedus2022switch}, and DeepSeekMoE~\cite{dai2024deepseekmoe}.
It has also been explored in sequence modeling tasks, as demonstrated by Time-MoE~\cite{shi2024time}, Moirai-MoE~\cite{liu2024moirai}, and Graph MoE~\cite{huang2025graph}.
However, MoE remains unexplored in next location prediction.  
To bridge this gap,  we introduce NextLocMoE, which  incorporates dual-level MoE modules—targeting location semantics and user behavioral patterns—paving the way for  MoE architecture in next location prediction.
For a more detailed discussion of MoE-related work,  please refer to App.~\ref{sec:related app moe}.
% !TEX root = neurips_2025.tex
\section{Problem Formulation}
Let $\mathcal{L} = \{loc_1, loc_2, \dots, loc_n\}$ be the set of locations, where each  $loc$ is  a triplet $(id, x, y)$, with $id$ being the location identifier and $(x, y)$ its spatial coordinates.
We define the temporal context set as $\mathcal{T} = \{(w, d)\ |\ w \in [0,6],\ d \in [0,23]\}$, where $w$ denotes  day-of-week and $d$ denotes time-of-day.
Let $\mathcal{D}_\nabla = \{\textit{dur}\}$ be the set of stay durations, indicating how long a user stays at a given location.

\textit{Definition 1 (\textbf{Record})} A record is defined as a tuple $s = (loc, (w, d), dur) \in \mathcal{L} \times \mathcal{T} \times \mathcal{D}_\nabla$, which indicates that a user visited location $loc$ for $dur$ hours at hour $d$ on day-of-week $w$.

\textit{Definition 2 (\textbf{Historical and Current Trajectory})} A user's mobility sequence can be partitioned into two disjoint segments: historical trajectory and current trajectory. 
The former is denoted as
$S_h = \{s_{t_1}, s_{t_1+1}, \dots, s_{t_1+M-1}\}$, which contains $M$  records used to model the user’s long-term behavioral preferences.
The latter is denoted as $S_c = \{s_{t_2}, s_{t_2+1}, \dots, s_{t_2+N-1}\}, (t_2 \geq t_1 + M)$, which includes the most recent $N$  records and is used to capture the user's short-term intent. Typically, $M > N$, ensuring that the historical trajectory spans a longer behavioral window.

\textit{Definition 3 (\textbf{Next Location Prediction})} Given a user's historical trajectory $S_h$ and current trajectory $S_c$, next location prediction aims to infer the identifier $id$ of the most likely next location $loc_{t_2 + N}$.
% !TEX root = neurips_2025.tex
\section{Methodology}

\begin{figure*}[h]
\centering
\includegraphics[scale=0.4]{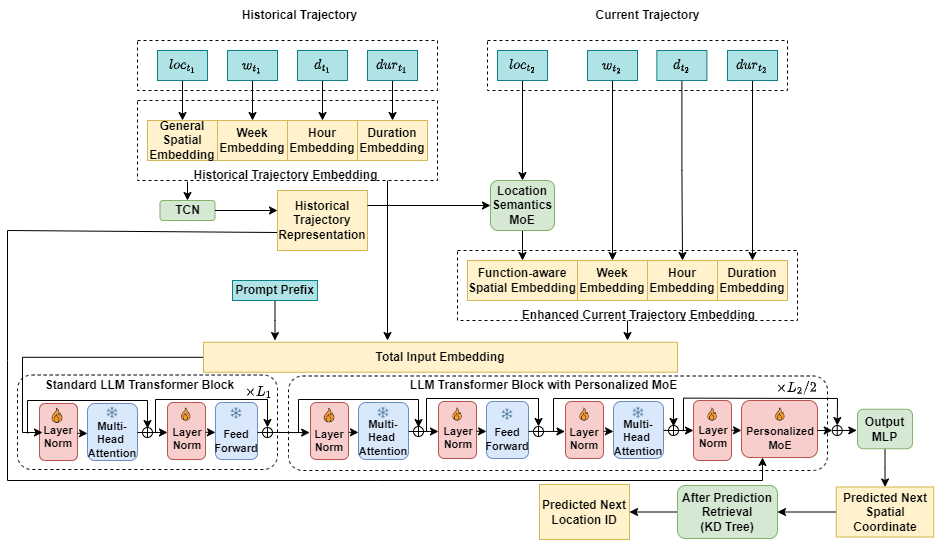}

\caption{Overall architecture of NextLocMoE, a Mixture-of-Experts enhanced LLM framework for next location prediction. 
It introduces a Location Semantics MoE to capture multi-functional spatial semantics (see Fig.~\ref{fig:MOE detail}(a), a Personalized MoE to model behavioral differences across user groups (see Fig.~\ref{fig:MOE detail}(a)), and a historical-aware router that incorporates long-term trajectory into expert routing.
\vspace{-15pt}}
\label{fig:model}
\end{figure*}

\subsection{Overall Architecture}
Fig.~\ref{fig:model} depicts the overall architecture of NextLocMoE. 
It takes  user’s historical and current trajectory as input, and encodes each record into spatial-temporal embedding (Sec.~\ref{sec:st emb}) by mapping coordinates $(x, y)$, temporal context ($w$ and $d$), and stay duration $dur$ to separate embeddings, which are then concatenated.
For current trajectory, we employ Location Semantics MoE (Sec.~\ref{sec:location MOE}) to  enrich spatial embedding with  location function semantics.
The function-aware spatial embedding is then combined with temporal embeddings to form the enhanced current trajectory embedding.
%In addition to contributing to prediction, the encoded historical trajectory is also provided as auxiliary input to the expert routing module to enhance its context-awareness during expert selection 
%The encoded historical trajectory supports both prediction and  expert routing (Sec.~\ref{sec:his router}). 
%For current trajectory, we incorporate Location Function MoE (Sec.~\ref{sec:location MOE}) to %adaptively 
%enrich location semantics. 
%It  encodes coordinates into a shared spatial embedding that captures general spatial properties, and  selects the top-$k$ most relevant function experts. 
%Each function expert encodes the same coordinates to a function-aware embedding, %and the aggregated expert outputs are combined with the shared embedding to produce an enhanced spatial vector.
%The resulting spatial vectors, along with time-related embeddings, are then concatenated as the enhanced current trajectory embedding.
%generates a shared spatial embedding for basic geographic attributes, and then  selects the top-$k$ most relevant function experts based on both current and historical context. 
%The aggregated outputs yield a refined spatial representation capturing fine-grained functional semantics.

Next, we concatenate historical trajectory embedding, enhanced current trajectory embedding, and a task-specific prompt (App.~\ref{fig:prompt}) to construct the total input embedding for LLM backbone (Sec.~\ref{sec:LLM and LoRA}).
Inspired by \cite{huang2024harder}, NextLocMoE employs only the first $L_1 + L_2$ layers of LLM:  $L_1$  standard LLM layers and subsequent $L_2$ layers augmented with Personalized MoE (Sec.~\ref{sec:persona MOE}) to model user behavioral patterns.
To improve expert routing robustness and reliability, NextLocMoE introduces a history-aware router (Sec.~\ref{sec:his router}) that incorporates long-term historical trajectories into expert selection.

To reduce parameter overhead, we fine-tune only the FFN sub-networks within MoE experts and all LayerNorm layers of the LLM, while freezing the remaining backbone layers. 
The final output of NextLocMoE is the predicted spatial coordinate of next location, obtained via an output MLP  head.
During inference, we apply a KD-Tree nearest neighbor search to map the predicted coordinate to a discrete location ID, completing the next location prediction task (Sec.~\ref{sec:taget}).

\subsection{Spatial-Temporal Embedding}
\label{sec:st emb}

In NextLocMoE, each component of a record is embedded through linear projection or embedding lookup.
Specifically,  spatial coordinates $(x, y)$ and stay duration $dur$ are normalized and  projected via linear layers to produce general spatial embedding $\mathbf{e}_{xy}$ and duration embedding $\mathbf{e}_{dur}$. 
For temporal context, $w$ and  $d$ are encoded via lookup tables, resulting in temporal embeddings $\mathbf{e}_w$ and $\mathbf{e}_d$.

For $S_h$, we concatenate the above four vectors along feature dimension to obtain  historical trajectory embedding $\mathbf{z}_h$, which %serves as input to the LLM backbone and expert router.
is used in two ways: as input to the LLM backbone and, after TCN encoding, as input to expert routers of both Location Semantics MoE and  Personalized MoE.
For each record in $S_c$, we adopt the same procedure to generate initial  embedding $\mathbf{e}_c^{(0)}$.
This embedding is used as input to  Location Semantics MoE, where it is combined with TCN-encoded historical trajectory representation  to guide expert selection and generate function-aware spatial embedding.

\begin{figure*}[h]
\centering
\includegraphics[scale=0.33]{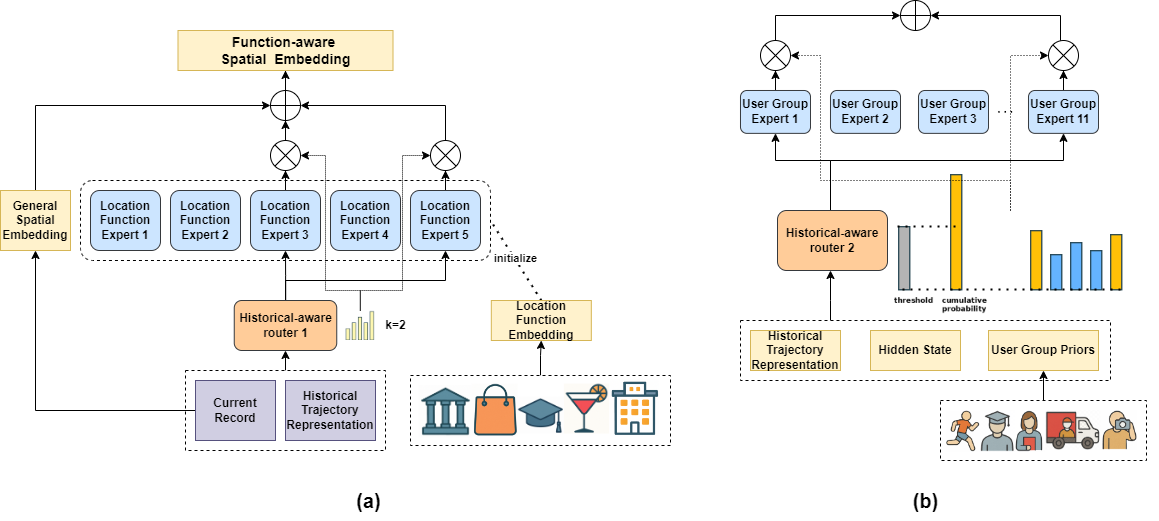}

\caption{Illustration of the two expert modules.
(a) Location Semantics MoE,
(b) Personalized MoE.}
\label{fig:MOE detail}
\vspace{-5pt}
\end{figure*}

\subsection{Location Semantics MoE}
\label{sec:location MOE}
In urban settings, a single location often serves multiple functions, such as shopping malls, schools, and public services.
Encoding such locations with multi-functional semantics  as a single vector limits the model’s expressiveness. 
To address this, NextLocMoE introduces Location Semantics MoE into current trajectory encoding (Fig.~\ref{fig:MOE detail}(a)), enabling fine-grained, function-aware location representations.

%This module takes as input the initial embedding of each record $\mathbf{e}_c^{(0)}$ in current trajectory and historical trajectory representation $\mathbf{h}^{\text{hist}}$.
This module takes as input the historical trajectory representation $\mathbf{h}^{\text{hist}}$ and the initial embedding of each record in the current trajectory, $\mathbf{e}_c^{(0)}$.
The former is obtained by encoding historical trajectory embedding $\mathbf{z}_h$ using a Temporal Convolutional Network (TCN)~\cite{lea2017temporal}:
\begin{equation}
    \mathbf{h}^{\text{hist}} = \mathrm{TCN}(\mathbf{z}_h).
\end{equation}
$\mathbf{e}_c^{(0)}$ and $\mathbf{h}^{\text{hist}}$ are fed into expert router to generate a scoring vector over $K_f$ function experts:
\begin{equation}
    \mathbf{r}^{\text{func}} = \mathrm{MLP}([\mathbf{e}_c^{(0)}; \mathbf{h}^{\text{hist}}]) \in \mathbb{R}^{K_f}.
\end{equation}
$\mathbf{r}^{\text{func}}$ is normalized via softmax to obtain expert selection probabilities $\mathbf{p}_i^{\text{func}}$.
The router then selects top-$k$ experts with  highest probabilities to enhance the semantic representation of the current record.

Each function expert $\mathbf{f}_i(\cdot)$ is a linear projection that maps spatial coordinates $(x, y)$ to a function-specific embedding. 
Its structure is identical to the mapping used for general spatial embedding $\mathbf{e}_{xy}$.
To promote interpretability and specialization, we predefine a set of location function categories (See App.~\ref{sec:loc prompt}) and encode their natural language descriptions using LLM.
The resulting embeddings are used to initialize the parameters of each expert, and these experts are subsequently fine-tuned during training.
This semantic initialization 
provides meaningful inductive biases, facilitating more effective semantics specialization and faster convergence.
%ensures that each expert starts with a distinct inductive bias, which improves specialization and training efficiency.

%Based on the above initialization and expert selection mechanisms, we further describe how to generate the function-aware spatial vector.
Given the selected top-$k$ function experts and their routing probabilities $\mathbf{p}_i^{\text{func}}$, we compute the summed location function specialized expert embedding as:
\begin{equation}
\mathbf{e}_{xy}^{\text{func}} = \sum_{i \in \mathrm{top}k(\mathbf{p}^{\text{func}})} \mathbf{p}_i^{\text{func}} \cdot \mathbf{f}_i(x, y).
\end{equation}
Following the design intuition of Deepseek-MoE\cite{dai2024deepseekmoe}, we treat general spatial embedding $\mathbf{e}_{xy}$ as a shared expert and combine it with $\mathbf{e}_{xy}^{\text{func}}$ to obtain the function-aware spatial embedding $\mathbf{e}_{xy}^{\text{enhanced}}$:
 \begin{equation}
\mathbf{e}_{xy}^{\text{enhanced}} = \mathbf{e}_{xy}+\mathbf{e}_{xy}^{\text{func}}.
\end{equation}
It is worth noting that Location Semantics MoE is applied only to the current trajectory records, not to the historical ones.
This is based on several considerations: (1) historical trajectories are used to model long-term behavioral patterns, where temporal dynamics outweigh fine-grained semantics;
(2) applying MoE to all trajectory records incurs high computational and memory costs; 
(3) function disambiguation is more important for current locations, whose semantics are directly tied to prediction.  
%Therefore, we limit the use of Location Function MoE to current trajectory records to strike a balance between modeling expressiveness and computational cost.

\subsection{Personalized MoE}
\label{sec:persona MOE}
To capture behavioral variations across user groups, NextLocMoE integrate Personalized MoE into the upper layers of the LLM backbone.
As shown in Fig.~\ref{fig:MOE detail}(b), we predefine $K_p$ prototypical user groups (see App.~\ref{sec:persona prompt}), each associated with a corresponding expert module.
For each user group, we encode its natural language description using an LLM. 
The resulting embeddings are then passed through a mean-pooling layer and a linear transformation to produce user group priors $\mathbf{e}^{\text{user}}_i,  (i=1,\cdots,K_p)$.
These priors provide semantic guidance and distinguish each expert by its behavioral identity.

Personalized MoE  receives the hidden state $\mathbf{x}$ from previous LLM layer, along with  historical trajectory representation $\mathbf{h}^{\text{hist}}$. 
For each expert $i$, we concatenate these with its prior:
\begin{equation}
\mathbf{z}_i^{\text{user}} = [\mathbf{x};\ \mathbf{h}^{\text{hist}};\ \mathbf{e}^{\text{user}}_i].
\end{equation}
$\mathbf{z}_i^{\text{user}}$ is first transformed by a multi-layer perceptron $\text{Fusion}(\cdot)$, and then projected by a linear gating function  $\text{Gate}(\cdot)$ to compute the relevance score $\mathbf{r}_i^{\text{user}}$:
\begin{equation}
\mathbf{r}_i^{\text{user}}=\mathrm{Gate}(\mathrm{Fusion}(\mathbf{z}_i^{\text{user}})).
\end{equation}
Stacking the scores across all experts yields the complete relevance vector:
\begin{equation}
    \mathbf{r}^{\text{user}} =\{\mathbf{r}_1^{\text{user}},\mathbf{r}_2^{\text{user}},\cdots,\mathbf{r}_{K_p}^{\text{user}}\}\in \mathbb{R}^{K_p}.
\end{equation}
We apply softmax over $\mathbf{r}^{\text{user}}$ to obtain the selection probability for each user group expert, $p_{i}^{\text{user}}$.

Unlike the fixed top-$k$ routing used in Location Semantics MoE, Personalized MoE adopts a confidence threshold-based expert routing strategy, inspired by \cite{huang2024harder}.
Specifically, we first sort experts by their selection probabilities $p_{i}^{\text{user}}$ and  sequentially select them until the cumulative probability exceeds a predefined threshold $\tau$:
\begin{equation}
\mathcal{E} = \{ i_1, i_2, \ldots, i_m \}, \quad \text{where} \quad \sum_{k=1}^{m} p_{i_k}^{\text{user}} \geq \tau.
\end{equation}
This design allows NextLocMoE to dynamically allocate expert capacity  based on the  complexity and ambiguity of user behavior.
For users with stable and clear mobility patterns, fewer experts are activated; for those with diverse or ambiguous behaviors, more experts are selected.
The activated experts perform feedforward transformations over the hidden state $\mathbf{x}$  and their outputs $\mathbf{h}_i$ are aggregated via weighted sum:
\begin{equation}
\mathbf{h}^{\text{out}} = \sum_{i \in \mathcal{E}} p_i^{\text{user}} \cdot \mathbf{h}_i.
\end{equation}

\subsection{Historical-aware Router}
\label{sec:his router}
Standard MoEs typically relies solely on the current input for expert selection~\cite{lepikhin2020gshard, fedus2022switch}.
However, relying only on current records for expert routing limits next-location prediction accuracy. 
Users with similar short-term routines may diverge in destination due to long-term behavior differences.
 For instance, after following the same morning routine from home to a metro station,  a student may go to university, while an office worker may head to a business district. 
Ignoring such historical context in expert routing would compromise both semantic modeling and personalized behavior modeling.

 To address this, NextLocMoE introduces a historical-aware router, which explicitly incorporates historical trajectories into expert selection.
  Specifically, we employ a TCN~\cite{lea2017temporal} to encode historical embeddings $\mathbf{z}_h$, yielding historical trajectory representation $\mathbf{h}^{\text{hist}}$, which is subsequently integrated into expert routing for both Location Semantics MoE and the Personalized MoE. 
 We choose TCN for its ability to efficiently capture long-range temporal dependencies and enable strong parallelization.
By incorporating historical trajectory representation, historical-aware router mitigates the  over-reliance on local context, stabilizes expert selection, and ultimately improves predictive accuracy and generalization.  

\begin{comment}
  During routing, we concatenate the current trajectory representation with the historical hidden state to form a context-aware input for expert scoring.
In Location Function MoE, the fused input is defined as $\mathbf{z}^{\text{func}} = [\mathbf{e}^{\text{cur}}; \mathbf{h}^{\text{hist}}]$. 
In Persona MoE, the input is further extended by incorporating each candidate expert’s semantic embedding: $\mathbf{z}_i = [\mathbf{x}; \mathbf{h}^{\text{hist}}; \mathbf{e}^{\text{persona}}_i]$.

The two MoE modules adopt different routing strategies. 
Location Function MoE uses a fixed Top-$k$ expert selection to capture diverse spatial semantics. 
In contrast, Persona MoE applies a confidence-thresholded dynamic routing strategy, allowing the number of activated experts to vary adaptively in response to user-specific behavioral diversity.
\end{comment}

\subsection{Streamlined LLM Backbone and Efficient Expert Adaptation}
\label{sec:LLM and LoRA}

To reduce computational overhead while maintaining predictive accuracy, NextLocMoE adopts a streamlined design.
Specifically, it retains only the first $L_1+L_2$ layers of LLM as its backbone.
The lower $L_1$ layers remain LLM layers, while the upper $L_2$ layers replace the original feedforward networks (FFNs) with  Personalized MoE. 
This design is inspired by \cite{skean2025layer}, which demonstrates that intermediate LLM representations are more stable and transferable than top-layer outputs, effectively filtering out high-entropy noise in downstream tasks. 
By truncating the model at intermediate layers, NextLocMoE preserves semantic encoding capacity while reducing architectural complexity.

To further limit trainable parameters and avoid overfitting, we freeze all multi-head attention modules and non-MoE FFNs in the LLM backbone, keeping only LayerNorm layers and  Personalized MoE experts trainable. 
Each user group expert is initialized using the parameters of the FFN it replaces, preserving representational continuity.
To enhance training efficiency, we apply Low-Rank Adaptation (LoRA) to  each user group expert.
LoRA introduces a small set of  trainable parameters in low-rank subspaces, allowing expert specialization and efficient personalization at minimal computational cost.

\subsection{Training Objective and Post-Prediction Retrieval}
\label{sec:taget}

The training objective of NextLocMoE combines a  regression loss and an expert entropy regularization term.
Given a batch of $B$ samples, NextLocMoE predicts the spatial coordinates $(\hat{x},\hat{y})$ of the next location.
The  regression loss is defined as the mean Euclidean distance to the ground truth $(x,y)$:
\begin{equation}
\mathcal{L}_{\text{dist}} = \frac{1}{B} \sum_{i=1}^{B} \left\| (\hat{x}_i, \hat{y}_i) - (x_i, y_i) \right\|_2.
\end{equation}
To encourage confident expert routing in Personalized  MoE and reduce unnecessary expert activation, we introduce an entropy regularization term:
\begin{equation}
\mathcal{L}_{\text{entropy}} = -\mathbb{E}_i \sum_j p_{i,j}^{\text{user}} \log p_{i,j}^{\text{user}}.
\end{equation}
The final training objective is a weighted combination of the two:
\begin{equation}
   \mathcal{L}_{\text{total}}= \mathcal{L}_{\text{dist}}+\lambda \times \mathcal{L}_{\text{entropy}},
\end{equation}
where $\lambda$ is a hyperparameter that balances the influence of the entropy regularization term.

During inference, to map the predicted coordinates to discrete location ID, NextLocMoE performs a KD-Tree-based nearest neighbor search. 
Specifically, a KD-Tree  is built from all candidate location coordinates, and the predicted coordinates are queried  to retrieve the ID of top-$k$ closest locations.

% !TEX root = neurips_2025.tex
\section{Experiment}
To evaluate the effectiveness of NextLocMoE, we conduct comprehensive experiments across three key dimensions: prediction accuracy, cross-city transferability, and interpretability.

\subsection{Experimental Setup}
\label{sec:setup}
\textbf{Datasets.} We conduct experiments on three real-world human mobility datasets: Kumamoto, Shanghai, and Singapore. Detailed descriptions are available in App.~\ref{sec:data desc}.

\textbf{Evaluation Metrics.} 
Following ~\cite{luo2021stan} and ~\cite{feng2018deepmove}, we adopt Hit@1, Hit@5, and Hit@10 for evaluation.

\textbf{Experiment Configuration.}
We follow  the user-level dataset partitioning strategy  in \cite{sun2021mfnp}, splitting users into training, validation, and test sets in a 7:1:2 ratio. 
%This ensures that test users do not appear during training. 
The historical and current trajectory lengths are set to $M=40$ and $N=5$. 
LLM backbone is LLaMA-3.2-3B, with $L_1=8$ and $L_2=4$.
%, with only the first 12 layers retained as the trajectory encoder. 
%Specifically, the first $L_1=8$ layers are standard Transformer blocks, while the next $L_2=4$ layers each replace every other FFN with Persona MoE. 
NextLocMoE is trained using the Adam optimizer and employs a ReduceLROnPlateau scheduler.
All experiments are conducted using mixed-precision training on four 32GB Tesla V100 GPUs.

\textbf{Baselines.}
We compare NextLocMoE with a wide range of baselines, including RNN-based models (GRU~\cite{chung2014empirical}, LSTM~\cite{graves2012long},  DeepMove~\cite{feng2018deepmove}); Transformer-based methods (MHSA~\cite{hong2023context}, CLLP~\cite{zhou2024cllp}, GETNext~\cite{yang2022getnext}, SEAGET~\cite{al2025seaget},  ROTAN~\cite{feng2024rotan}), and LLM-based approaches (LLM4POI~\cite{li2024large}, NextLocLLM~\cite{liu2024nextlocllm}, Llama-Mob~\cite{tang2024instruction}, LLMMob~\cite{wang2023would},  ZSNL~\cite{beneduce2025large}).
Details are available in App.~\ref{sec:baseline intro}

\subsection{Experimental Result}
\subsubsection{Fully-supervised Prediction Comparison}

Table.~\ref{tab:fs result} presents the fully-supervised next location prediction results on all three datasets.
RNN-based models  perform poorly, indicating that local temporal modeling is insufficient to capture the complex spatiotemporal  mobility patterns.
Methods like CLLP, GETNext, SEAGET, ROTAN, and LLM4POI rely on user IDs or user embeddings, which fail to generalize under user-level partitioning where test users are unseen during training.
Llama-Mob, the winner of  2024 HuMob Challenge, performs better but remains limited in modeling multi-functional location semantics and user behavioral patterns.
In contrast, NextLocMoE introduces two  innovations: Location Semantics MoE for fine-grained semantic modeling of locations, and Personalized MoE for user behavioral patterns.
These designs lead to consistent state-of-the-art performance across all datasets and metrics.

\begin{table}[!t]
  \caption{Fully-supervised next location prediction results.}
  \label{tab:fs result}
   \resizebox{\textwidth}{!}{
 \begin{tabular}{l|ccc|ccc|ccc}
    \toprule
    {Method } &  \multicolumn{3}{c|}{Kumamoto}   &  \multicolumn{3}{c|}{Shanghai}  &  \multicolumn{3}{c}{Sinapore}  \\
    & Hit@1 & Hit@5 & Hit@10  & Hit@1 & Hit@5 & Hit@10  & Hit@1 & Hit@5 & Hit@10 \\ 
    \cmidrule(r){1-1}\cmidrule(r){2-4}\cmidrule(r){5-7}\cmidrule(r){8-10}
    GRU &3.213\% & 6.720\% & 8.735\% & 19.69\% & 25.90\% & 29.04\% & 2.682\% & 6.051\% & 7.784\%\\
    LSTM & 3.192\% & 6.483\% & 8.514\% & 22.03\% & 28.81\% & 31.33\% & 3.197\% & 8.698\% & 10.46\% \\
    MHSA & 2.982\% & 9.203\% & 11.77\% & 48.40\% & 56.62\% & 62.21\% & 4.874\% & 13.54\% & 19.38\%\\
    DeepMove & 11.11\% & 20.71\% & 24.46\% & 53.48\% & 62.13\% & 67.70\% & 6.650\% & 20.00\% & 31.08\%\\
    GetNext & 12.68\% & 24.57\% & 29.80\% & 55.18\% & 64.17\% & 71.17\% & 6.498\% & 25.80\% & 32.04\%\\
    CLLP & 10.69\% & 17.79\% & 21.96\% & 56.24\% & 65.39\% & 72.08\% & 7.712\% & 26.98\% & 34.99\% \\
    SEAGET & 12.79\% & 24.66\% & 29.99\% & 55.39\% & 65.12\% & 70.93\% & 6.512\% & 25.94\% & 32.56\%\\
    NextLocLLM & 13.57\% & 24.78\% & 31.16\% & 59.62\% & 66.93\% & 72.81\% & 7.823\% & 30.64\% & 36.15\%\\
    ROTAN & 13.01\% & 26.19\% & 32.37\% & 57.92\% & 66.83\% & 72.06\% & 6.892\% & 27.71\% & 35.56\%\\
    LLM4POI & 13.17\% & 26.88\% & 30.11\% &58.83\% & 67.72\% & 72.47\% & 7.952\% & 31.69\% & 38.88\%\\
    SoloPath & 13.75\% & 27.80\% & 34.61\% & 60.21\% & 67.92\% & 69.24\% & 8.102\% & 30.00\% & 37.51\% \\
    Llama-Mob & \underline{15.78\%} & \underline{33.55\%} & \underline{43.42\%} & \underline{61.81\%} & \underline{69.36\%} & \underline{73.45\%} & \underline{8.577\%} & \underline{32.17\%} & \underline{41.21\%}\\
    LLMMob & 10.95\% & 25.54\% & 35.77\% &51.17\% & 60.93\% & 63.31\% & 6.933\% & 21.07\% & 30.70\%\\
    ZS-NL & 8.811\%  & 22.97\% & 31.76\% & 39.92\% & 47.71\% & 50.98\% & 4.199\% & 14.68\% & 20.11\%\\
     \cmidrule(r){1-1}\cmidrule(r){2-4}\cmidrule(r){5-7}\cmidrule(r){8-10}
     NextLocMoE & \textbf{17.77\%} & \textbf{39.19\%} & \textbf{50.28\%} & \textbf{64.93\%} & \textbf{75.88\%} & \textbf{77.43\%} & \textbf{9.733\%} & \textbf{34.34\%} & \textbf{40.71\%} \\ 
    \cmidrule(r){1-1}\cmidrule(r){2-4}\cmidrule(r){5-7}\cmidrule(r){8-10}
\end{tabular}}
\vspace{-5pt}
\end{table}

\subsubsection{Zero-shot  Prediction Comparison}

To assess generalization, we conduct zero-shot experiments on Kumamoto dataset, where the model is directly tested using parameters trained on other cities without any fine-tuning.
Since location IDs differ across cities, ID-based non-LLM models are unable to make effective predictions in this setting. 
Thus, we compare only transferable methods:  Llama-Mob, LLMMob, ZS-NL, NextLocLLM, and our proposed NextLocMoE.
As shown in Table~\ref{tab:zero-shot}, NextLocMoE achieves the best across all metrics.
We attribute this to its explicit modelling of location semantics and user behavior: the Location Semantics MoE encodes functional semantics agnostic to location IDs, while Personalized MoE adapts to user behavior through role-based experts—enabling robust transfer to unseen cities.

\subsubsection{Inference Time}
%Inference speed is also a key factor. 
We report the total inference time of the transferable LLM-based models on Kumamoto test set in Table~\ref{tab:inference-time}.
Llama-Mob relies on locally deployed LLM with separate prompt construction and autoregressive decoding per trajectory, resulting in highly sequential and time-consuming inference. 
%where a prompt must be constructed and processed separately for each trajectory. 
%The generation process is fully autoregressive and highly sequential, resulting in extremely long inference times.
LLMMob, offloading computation via external APIs, still suffers from serialized generation.
%Although LLMMob offloads computation by calling  external API, thereby alleviating the local deployment burden, its generative inference remains constrained by prompt construction and serialized decoding.
%In contrast, NextLocMOE adopts a unified neural architecture for vectorized modeling and supports fully parallel inference. 
%Thanks to this design, the model can fully leverage GPU acceleration and execute inference efficiently in batch mode. 
NextLocMoE adopts a unified architecture that supports batch inference and GPU parallelism.
It completes inference in 268 seconds—a 600× speedup over Llama-Mob and 120× faster than LLMMob.
%Experimental results show that NextLocMOE completes inference on the entire Kumamoto test set in just 268 seconds, achieving a ~600× speedup over Llama-Mob and more than 120× faster inference than LLMMob.

\begin{table}[!t]
  \centering
  \begin{minipage}{0.58\textwidth}
    \caption{Zero-shot  Prediction Result (Kumamoto).}
    \label{tab:zero-shot}
    \setlength{\tabcolsep}{4pt}
    \begin{tabular}{l|l|l|l}
      \toprule
      Method & Hit@1 & Hit@5 & Hit@10 \\
      \midrule
      Llama-Mob & 15.78\% & 33.55\% & 43.42\% \\
      LLMMob & 10.95\% & 25.54\% & 35.77\% \\
      ZS-NL & 8.811\% & 22.97\% & 31.76\% \\
      NextlocLLM(Shanghai$\rightarrow$) & 13.14\% & 28.68\% & 39.26\% \\
      NextlocLLM(Singapore$\rightarrow$) & 11.73\% & 26.95\% & 37.53\% \\
      \midrule
      NextLocMoE(Shanghai$\rightarrow$) & 16.02\% & 36.06\% & 48.42\% \\
      NextLocMoE(Singapore$\rightarrow$) & 15.81\% & 34.66\% & 47.41\% \\
      \bottomrule
    \end{tabular}
  \end{minipage}%
  \hfill
  \begin{minipage}{0.38\textwidth}
    \caption{Inference Time (Kumamoto).}
    \label{tab:inference-time}
    \setlength{\tabcolsep}{6pt}
    \begin{tabular}{l|c}
      \toprule
      Method & Time (s) \\
      \midrule
      Llama-Mob & 158688 \\
      LLMMob & 33408 \\
      NextLocMoE & 268 \\
      \bottomrule
    \end{tabular}
  \end{minipage}
  \vspace{-15pt}
\end{table}

\subsubsection{Case Study}

We analyze two representative trajectories from  Singapore dataset (Fig.~\ref{fig: case}).
Though their current trajectory exhibit similar spatial patterns, their  historical trajectories differ: the first is centered around academic zones, while the second frequently appears in commercial and tourist areas.
%However, they differ significantly in their long-term history trajectories:  the first trajectory is predominantly associated with universities and academic areas, while the second frequently occurs near commercial zones and tourist attractions.
In Location Semantics MoE, current location in the first trajectory is assigned higher weights to Education and Entertainment, whereas that in the second trajectory favors Entertainment and Commercial.
In Personalized MoE, the first user is routed to student and teacher experts, whereas the second strongly activates the tourist expert.
Ultimately,  NextLocMoE produces distinct and correct next location predictions for the two cases—highlighting its ability to make interpretable and effective forecasts.

%This demonstrates that,  NextLocMoE can accurately perceive user roles based on historical behavioral differences.
%Ultimately, NextLocMOE assigns distinct next-location predictions to the two trajectories—despite their surface-level similarity in recent behavior—based on expert allocation grounded in both functional semantics and user persona. Both predictions successfully match the ground-truth next location, illustrating the model’s ability to go beyond short-term cues and leverage historical context through expert routing for personalized next-location forecasting.

\begin{comment}
\begin{figure*}[h]
\centering
\includegraphics[scale=0.28]{figure/case_study.drawio.png}

\caption{Case study for NextLocMoE.}
\label{fig: case}
\vspace{-10pt}
\end{figure*}
\end{comment}
\subsubsection{Hyperparameter Sensitivity}
\begin{figure*}[h]
\centering
\includegraphics[scale=0.25]{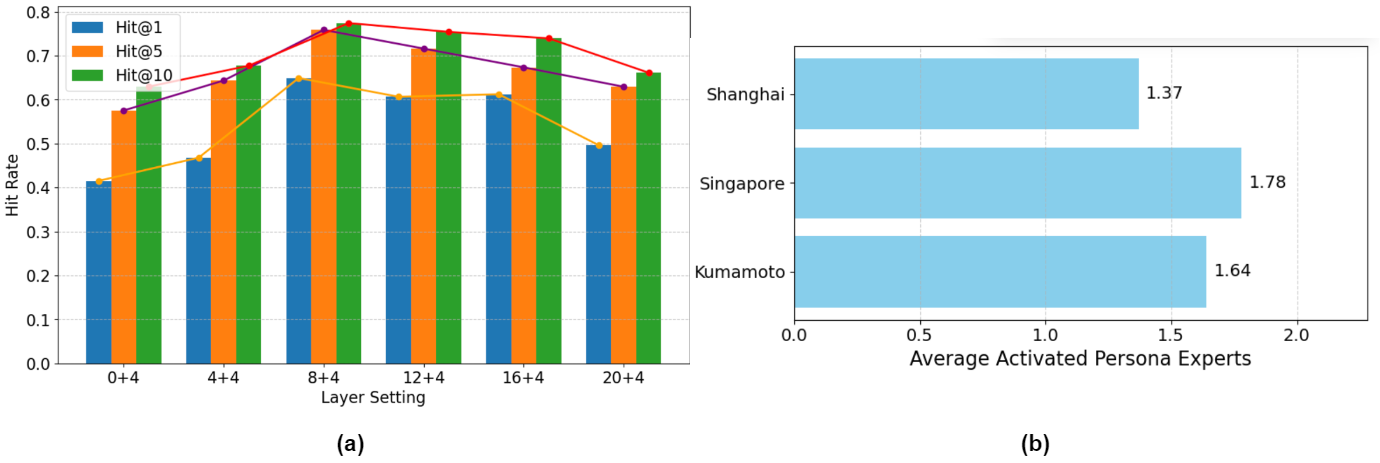}

\caption{(a) Hyperparameter  sensitivity; (b) Personalized expert activation analysis}
\label{fig: ex}
\vspace{-10pt}
\end{figure*}

We examine how freezing different numbers of LLM layers affects performance, while keeping the 4 layers integrated with Personalized MoE.
%To investigate how the number of retained LLM layers affects the performance of NextLocMOE, we conduct experiments by varying the number of frozen lower layers while keeping the final 4 layers integrated with the Persona MoE structure.
As shown in Fig.~\ref{fig: ex}(a), freezing 8 layers yields the best results.
Fewer frozen layers  lead to poorer generalization, while freezing more than 8 layers degrades performance. 
%In contrast, when fewer layers are frozen (e.g., 0 or 4), the performance drops, suggesting that shallow representations generalize poorly.
%As the number of frozen layers increases beyond 8 (e.g., 12 or 16), performance gradually degrades, and it significantly deteriorates when freezing reaches 20 layers.
%This trend indicates that intermediate layers of decoder-only LLMs offer stronger adaptability, which aligns with 
This supports prior findings \cite{skean2025layer} that intermediate layers in decoder-only LLMs offer stronger adaptability.
Based on this trade-off, we adopt the 8-layer freezing configuration as default.
%Considering both predictive performance and training cost, we adopt the 8 frozen layers configuration as the default setting for NextLocMOE.

\subsubsection{Personalized Expert Activation Analysis}
We analyze the average number of activated experts in Personalized MoE  (Fig.~\ref{fig: ex}(b)).
NextLocMoE activates 1.37 experts on Shanghai, 1.78 on Singapore, and 1.64 on Kumamoto—consistently fewer than 2. 
This shows NextLocMoE’s ability to adaptively engage a minimal set of user group experts based on user behavior complexity, ensuring personalized modeling with low inference cost.
%, the average number of experts activated by NextLocMOE is 1.37 on Shanghai, 1.78 on Singapore, and 1.64 on Kumamoto, all of which are less than 2.
%This result indicates that the model can adaptively activate only a small subset of critical persona experts based on the complexity of user behavior patterns. Such selective activation enables personalized modeling with minimal computational overhead, striking a balance between performance and inference efficiency.
% !TEX root = neurips_2025.tex
\section{Conclusion}
We propose NextLocMoE, a dual-level Mixture-of-Experts (MoE) enhanced large language model for next location prediction. 
It incorporates two complementary modules: Location  MoE, which captures fine-grained location functional semantics using a fixed top-$k$ expert routing, and Personalized MoE, which models user behavioral patterns diversity via confidence-thresholded dynamic routing.
%and —to jointly model the multi-functional semantics of geographic locations and the behavioral heterogeneity across user groups.
%Location Function MoE dynamically enhances spatial representations by fusing a general spatial embedding with function-specific experts initialized from textual descriptions, enabling fine-grained modeling of location semantics.
%Persona MoE, on the other hand, leverages user role descriptions and historical behavior features to activate experts through a confidence-thresholded dynamic routing mechanism for personalized modeling.
To improve contextual awareness and reliability in expert selection, NextLocMoE introduces a historical-aware router, which explicitly incorporates long-term historical trajectories during expert routing. 
%This design improves both the stability and rationality of expert assignment decisions.
Empirical results %on multiple real-world urban mobility datasets, including zero-shot evaluation settings, 
show that NextLocMoE outperforms existing baselines in accuracy, generalization, and inference speed. 
Case study also shows its interpretability.
Nonetheless, NextLocMoE incurs notable training-time memory costs due to maintaining full FFN sub-networks per user group expert.
Future work will explore expert compression techniques, such as weight-splitting from Llama-MoE~\cite{zhu2024llama}, to reduce this overhead.

\small
\newpage
\bibliographystyle{unsrt}
\bibliography{reference}

\appendix
% !TEX root = neurips_2025.tex
\newpage
\section{Detailed Related Work}

\subsection{Next Location Prediction}
\label{sec:related app next loc}
Next Location prediction aims to forecast the most probable place a user will visit in the near future, based on his/her past mobility trajectory. 
This task has attracted increasing research interest.
 Over time, models have evolved significantly to better capture the complex temporal dynamics, spatial semantics, and behavioral diversity inherent in human mobility~\cite{chekol2022survey, rajule2023mobility,zhang2018mobility}.
 Broadly, existing methods can be categorized into three major paradigms: RNN-based models that emphasize sequential learning~\cite{sherstinsky2020fundamentals}, attention-based models that enhance long-range context integration~\cite{vaswani2017attention}, and LLM-based models that leverage pretrained knowledge and reasoning capabilities~\cite{achiam2023gpt,zhu2024llama}. 
 Below, we provide a detailed review of representative methods within each category.

\subsubsection{RNN-based Next Location}
Early approaches to next-location prediction primarily relied on recurrent neural networks, such as GRU~\cite{chung2014empirical} and LSTM~\cite{graves2012long}, to model sequential dependencies. 
DeepMove~\cite{feng2018deepmove}  jointly models short-term interests and long-term preferences, capturing user mobility patterns over multiple timescales.
SASRM~\cite{zhang2020sasrm} introduces a semantic- and attention-enhanced spatio-temporal recurrent model, which better captures location semantics and contextual dependencies. 
MCN4Rec~\cite{li2024mcn4rec} takes a multi-perspective approach, collaboratively learning from both local and global views to model heterogeneous relationships among users, POIs, temporal factors, and activity types.  
\cite{zhang2022beyond} extend the theoretical foundation of mobility prediction by introducing a new upper bound that incorporates not only sequential patterns but also contextual features such as time and location categories.

In parallel, several models address practical challenges like data privacy and label scarcity. 
SecureDeepMove~\cite{liu2024secdm} integrates secret sharing and secure two-party computation to perform inference without compromising user privacy.
SelfMove~\cite{hong2023mobility} adopts a self-supervised learning strategy to disentangle time-invariant and time-varying factors, enabling training without labeled next-POI data. 
\cite{hasan2022lstm} design an LSTM-based system that effectively leverages sequential and temporal cues from device-level mobility logs.

Hybrid architectures also emerge.
SAB-GNN~\cite{xue2022multiwave} fuses LSTM with a Graph Neural Network to jointly capture spatial dependencies across urban regions and temporal dynamics from user mobility and web search activity. 
Notably, it incorporates decaying public awareness signals to forecast multiwave patterns in mobility—demonstrating the flexibility of RNN-based frameworks in complex real-world scenarios.

\subsubsection{Attention-based Next Location Prediction}

With the rise of the Transformer architecture~\cite{vaswani2017attention}, attention-based methods have rapidly become the mainstream in next-location prediction due to their superior ability to model long-range dependencies and capture complex spatial-temporal interactions. 
These models often extend attention mechanisms with auxiliary data, personalized encodings, or graph structures to enhance predictive performance and generalization.

Several works enhance spatial-temporal reasoning via graph-augmented attention.
TrajGraph~\cite{zhao2024trajgraph} employs a graph Transformer to efficiently encode spatiotemporal context under reduced computational complexity.
GETNext~\cite{yang2022getnext} and SEAGET~\cite{al2025seaget} construct trajectory flow graphs to incorporate collaborative mobility signals into attention-based models.
AGCL~\cite{rao2024next} introduces a multi-graph learning framework with adaptive POI graphs, spatial-temporal attention, and bias correction. 
iPCM~\cite{song2025integrating} combines global trajectory data with personalized user embeddings using a Transformer encoder and probabilistic correction module.

Another line of work explores behavior modeling and user preference learning.
MHSA~\cite{hong2023context} models transition relations among locations using multi-head self-attention. 
CLLP~\cite{zhou2024cllp} fuses local and global spatiotemporal contexts to track evolving user interests.
CTLE~\cite{lin2021pre} maps contextual encodings into a target location embedding, followed by bidirectional Transformer modeling.
MCLP~\cite{sun2024going} leverages topic models to extract latent user preferences and enhances arrival time estimation via attention.
FHCRec~\cite{chen2025enhancing} captures both long- and short-term patterns through hierarchical contrastive learning over subsequences.
STMGCL~\cite{jia2023improving} introduces temporal group contrastive learning within a self-attention encoder to uncover user preference groups.

Auxiliary signals are widely integrated.
PRPPA~\cite{liang2019deep} combines static user profiles, recent check-in behavior, and temporal point processes into a unified attention framework.
SanMove~\cite{wang2023sanmove} proposes a non-invasive self-attention module that utilizes auxiliary trajectory signals to learn short-term preferences. 
TCSA-Net~\cite{sun2022tcsa} jointly captures long- and short-term mobility patterns from sparse and irregular trajectories. LoTNext~\cite{xu2024taming} addresses the long-tail challenge via graph and loss adjustments that rebalance POI interaction distributions.

Domain-specific and event-aware attention models have also emerged.
Physics-ST~\cite{gao2024physics} infuses physics priors into human mobility modeling by formulating movement as governed by potential energy dynamics, combined with graph-based attention and temporal correction. 
~\cite{wang2023towards}  incorporates event embeddings to represent both routine behaviors and disruptions.
The BERT-based method of \cite{terashima2023human} repurposes pretrained language encoders for trajectory modeling.
~\cite{shukla2024exploring} uses an encoder–decoder attention structure for coordinate-level  prediction.

\subsubsection{LLM-based Next Location Predction}
In recent years, breakthroughs in large language models (LLMs)\cite{achiam2023gpt,liu2024deepseek,touvron2023llama} have sparked growing interest in their application to next-location prediction. 
These models offer strong reasoning abilities, contextual understanding, and pre-trained world knowledge that can complement traditional mobility modeling frameworks.
Llama-Mob\cite{tang2024instruction} and LLMMob~\cite{wang2023would} incorporate task-specific prompting strategies to adapt LLMs for spatial prediction tasks.
Going further, NextLocLLM~\cite{liu2024nextlocllm} introduces a dual-role usage of LLMs, functioning as both semantic enhancer and next-location predictor, thereby improving both accuracy and generalization across mobility datasets.
AgentMove~\cite{feng2025agentmove} decomposes the next-location prediction task into three specialized components: a spatial-temporal memory module that captures individual behavioral patterns, a world knowledge generator that infers structural and urban influences, and a collective knowledge extractor that models shared mobility patterns across populations. 
Meanwhile, CausalMob~\cite{yang2024causalmob} introduces a causality-inspired framework that leverages LLMs to extract latent intention signals tied to external events. It then estimates their causal effects on user mobility while controlling for spatial and temporal confounders—highlighting the potential of LLMs to go beyond pattern recognition and engage in causal reasoning within human mobility modeling.

\subsection{Mixture of Experts}
\label{sec:related app moe}
 Mixture-of-Experts (MoE) has become a foundational approach for scaling large models while maintaining computational efficiency. 
 Unlike dense models that activate all parameters for every input, MoE architectures route each token or input to a small subset of specialized experts, drastically reducing the per-example computation~\cite{lo2024closer}.
Early works such as GShard \cite{lepikhin2020gshard} and Switch Transformer \cite{fedus2022switch} pioneered this direction.
GShard introduced a scalable training framework with automatic sharding support, enabling a 600B-parameter Transformer to be trained on 2048 TPUs. Switch Transformer further simplified the routing mechanism by activating only one expert per token, leading to better training stability and communication efficiency, and achieving 7× speedups during pretraining. These foundational designs demonstrate the practicality of scaling models to the trillion-parameter regime without linearly increasing computational cost.

Subsequent works have focused on improving expert specialization, routing flexibility, and deployment efficiency. DeepSeekMoE \cite{dai2024deepseekmoe} introduces fine-grained expert segmentation and shared experts to encourage non-overlapping expertise and reduce redundancy. 
PMoE \cite{jung2024pmoe} adopts an asymmetric transformer layout, with shallow layers handling general knowledge and deep layers using progressively added experts for continual learning, mitigating catastrophic forgetting. 

Beyond training from scratch, several methods propose transforming existing dense models into MoE architectures.
LLaMA-MoE \cite{zhu2024llama} partitions feed-forward layers of LLaMA-2 and uses continual pretraining to preserve language capability while introducing sparse expert routing. 
MoE Jetpack \cite{zhu2024moe} repurposes dense model checkpoints and introducing a hyperspherical adaptive MoE layer for efficient fine-tuning.

Efficiency during inference and dynamic routing has also been actively explored.
\cite{huang2024harder} adjusts the number of active experts per input based on difficulty, dispatching more experts for complex reasoning tasks.
\cite{lu2024not} propose post-training strategies to reduce active parameters per task, improving MoE deployability without retraining. 
MixLoRA \cite{shen2024multimodal} adapts MoE to multimodal instruction tuning by constructing instance-specific low-rank LoRA adapters to reduce task interference.

\section{Location Function Natural Language Description}
\label{sec:loc prompt}

We define a set of location function categories that capture the diverse functional roles a location may serve.
Table~\ref{tab:loc desc} provides natural language descriptions of these  predefined semantic categories. 
Each category reflects a distinct aspect of urban space usage and is used to initialize corresponding experts with LLM-encoded semantic priors.

\begin{table}[!t]
  \caption{Location Function Natural Language Description.}
  \label{tab:loc desc}
   \setlength{\tabcolsep}{4pt}
 \begin{tabular}{l|p{11cm}}
 \toprule
Location Function & Description \\
 \midrule
 Entertainment & This category includes scenic spots, sports venues, and recreational facilities, offering activities for leisure, entertainment, and social interactions.Typical examples include amusement parks, cinemas, stadiums, and bars. Users often visit for relaxation, nightlife, sports, and cultural experiences, with peak times in evenings and weekends.\\
  \midrule
 Commercial & This category encompasses businesses, financial institutions, automotive services, shopping centers, and dining establishments, supporting daily consumer and professional needs. Typical examples include malls, banks, car dealerships, and restaurants. Users often visit during working hours or weekends for shopping, financial transactions, or dining.\\
  \midrule
 Education & This category covers institutions focused on academic, cultural, and scientific learning. Typical examples include schools, universities, libraries, and research centers. Users often visit on weekdays for study, teaching, research, and cultural enrichment. \\
  \midrule
 Public Service & This category includes government offices, healthcare facilities, transportation hubs, and other essential public infrastructure. Typical examples include city halls, hospitals, bus stations, and utility centers. Users often visit for administrative tasks, medical needs, commuting, or essential services, with varied peak hours depending on the service type.\\
  \midrule
 Residential & This category comprises housing areas, mixed-use developments, and temporary accommodations. Typical examples include apartment complexes, residential neighborhoods, and hotels. Users often visit for long stays, typically peaking in the evenings, weekends, and holidays.\\
 \bottomrule
\end{tabular}
\end{table}

\section{User Group Natural Language Description}
\label{sec:persona prompt}

We define a set of representative user groups based on common mobility behaviors. 
These groups reflect distinct travel patterns across different roles in urban settings, similar to those introduced in~\cite{jiawei2024large}. Table~\ref{tab:persona desc} provides natural language descriptions for the predefined user groups.

\begin{table}[!t]
  \caption{User Group Natural Language Description.}
  \label{tab:persona desc}
   \setlength{\tabcolsep}{4pt}
 \begin{tabular}{l|p{11cm}}
 \toprule
 User Group & Description \\
 \midrule
 Student & This persona represents individuals who typically travel to and from educational institutions at regular times, such as morning arrivals and afternoon departures. Their mobility is highly time-structured and centered around campuses, libraries, and nearby service areas.\\
 \midrule
 Teacher & This persona regularly commutes to educational institutions during weekday mornings and returns home in the late afternoon or early evening. Their travel patterns align closely with school schedules, often involving brief visits to nearby commercial or service areas.\\
 \midrule
 Office Worker & This persona has a fixed daily commute, traveling to office districts or commercial centers in the morning and returning home in the evening. Their mobility follows a consistent weekday routine with limited variation.\\
 \midrule 
 Visitor & This persona tends to travel throughout the day with less predictable patterns. They frequently visit tourist attractions, cultural landmarks, dining areas, and shopping districts, especially in central urban zones.\\
 \midrule 
 Night Shift Worker & This persona often travels outside of standard business hours, especially during late evenings or at night. Common destinations include hospitals, factories, 24-hour service locations, and late-night dining spots. \\
 \midrule
 Remote Worker & This persona has non-standard travel patterns. They frequently visit coworking spaces, cafÃ©s, or quiet public environments at various hours of the day, with flexible scheduling that may shift across weekdays.\\
 \midrule
 Service Industry Worker & This persona has irregular travel times throughout the day. They frequently move between restaurants, shopping areas, entertainment venues, and other customer-facing POIs, reflecting shift-based work in dynamic urban zones.\\
 \midrule
 Public Service Official & This persona often works in rotating shifts, leading to variable travel patterns across different times of the day and night. Common destinations include government offices, transport hubs, hospitals, and administrative centers.\\
 \midrule
 Fitness Enthusiast & This persona is active during early mornings, evenings, or weekends. Their mobility revolves around gyms, sports facilities, parks, and wellness-related POIs. Visit durations tend to be regular and intentional.\\
 \midrule 
 Retail Employee & This persona typically begins travel in the late morning and returns in the evening. Their destination patterns focus on malls, retail stores, and service clusters, reflecting the opening and closing hours of retail operations.\\
 \midrule
 Undefined Persona & This persona does not clearly belong to any predefined behavioral category. Their travel patterns may be irregular, spontaneous, or inconsistent across time and location.\\
  \bottomrule
\end{tabular}
\end{table}

\section{Prompt Prefix}
\begin{figure*}[h]
\centering
\includegraphics[scale=0.35]{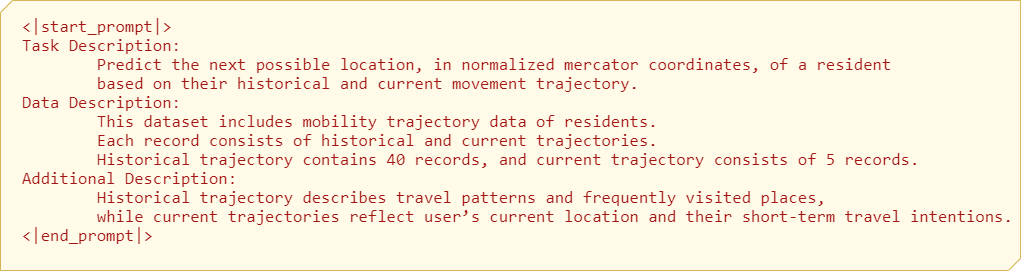}

\caption{Prompt prefix used in NextLocMoE.}
\label{fig:prompt}
\end{figure*}

Fig.\ref{fig:prompt} outlines the specific task and data prompt prefix used in NextLocMoE. 
The prompt prefix begins by defining the task and providing a detailed description of the dataset structure.
 Additionally, the Additional Description section emphasizes how to think about this task using the provided data.

\section{Dataset Description}
\label{sec:data desc}
We use three real-world mobility datasets to validate the effectiveness of NextLocLoE, and the detailed descriptions of these datasets are as follows:

\textbf{Kumamoto} \footnote{\url{https://zenodo.org/records/13237029}} This is an open-source and anonymized dataset of human mobility trajectories from mobile phone location data provided by Yahoo Japan Corporation. The location pings are discretized into 500meters $\times$ 500meters grid cells and the timestamps are rounded up into 30-minute bins. 

\textbf{Shanghai} \footnote{\url{https://github.com/vonfeng/DPLink}} This dataset contains mobility records that cover the metropolitan area of Shanghai from April 19 to April 26 in 2016. 
We selected the core areas of Puxi and the neighborhoods within the Middle Ring Road of Pudong.
The location pings are discretized into 200meters $\times$ 200meters grid cells.

\textbf{Singapore} This data is collected by one mobile SIM card company in Singapore. It is proprietary and provided under a restricted research agreement with the data owner. We choose the locations in central Singapore. The location pings are discretized into 200meters $\times$ 200meters grid cells.

\begin{table}[!t]
  \caption{Dataset Description.}
  \label{tab:dataset}
   \setlength{\tabcolsep}{4pt}
 \begin{tabular}{l|c|c|c|c|c}
 \toprule
 Dataset & Num of Records & Time Span (day) & Num of Users & Avg Interval (min) & Num of Locations \\
 \midrule
 Kumamoto & 6696506 & 60  & 17965 & 68.4 & 6000 \\
 Shanghai & 1337256 & 8  & 30421 & 94.8 & 10085 \\
 Singapore &  2714672 & 31 & 17098 & 61.4 & 4720\\
 \bottomrule
\end{tabular}
\end{table}

\section{Baseline Description}
\label{sec:baseline intro}
The details of baseline methods are briefly summarized as follows. 
\begin{itemize}
    \item LSTM~\cite{graves2012long} A type of recurrent neural network capable of learning
 order dependence in sequence prediction problems.
    \item GRU~\cite{chung2014empirical} Similar to LSTMs, GRUs are a streamlined version that use gating
 mechanisms to control the flow of information and are effective in sequence modeling tasks.
    \item DeepMove~\cite{feng2018deepmove} This model uses the attention mechanism  to
 combine historical trajectories with current trajectories for prediction.
    \item MHSA~\cite{hong2023context} An attention-based model that integrates various contextual
 information from raw location visit sequences.
 \item CLLP~\cite{zhou2024cllp} It integrates both local and global spatiotemporal contexts to better capture dynamic user interests.
 \item GETNext~\cite{yang2022getnext} It introduces global trajectory flow graphs and graph-enhanced Transformer models.
 \item SEAGET~\cite{al2025seaget} It uses graph Transformer to leverage collaborative mobility signals to improve predictive performance.
 \item ROTAN~\cite{feng2024rotan} It proposes a brand new Time2Rotation technique to capture the temporal information.
 \item LLM4POI~\cite{li2024large} It effectively uses the abundant contextual information present in LBSN data.
 \item Llama-Mob~\cite{tang2024instruction} It instruction tuned Llama for mobility prediction. For alignment, we replace its backbone to Llama3.2-3B, as NextLocMoE uses.
 \item NextLocLLM~\cite{liu2024nextlocllm} It leverages  LLM as both a semantic enhancer and a predictor.
 \item LLmMob~\cite{wang2023would} It introduces concepts
 of historical and contextual stays to capture the long-term and short-term dependencies in
 human mobility.
 \item ZSNL~\cite{beneduce2025large} It is a  purely prompt based model designed for zero-shot
 next location prediction.
\end{itemize}

\section{Further Hyperparameter Settings}
\label{sec:hyper}
We provide the full hyperparameter list for Kumamoto dataset in Table~\ref{tab:kuma hyper}.

\begin{table*}[!t]
  \caption{Hyperparameter list for Kumamoto dataset.}
  \label{tab:kuma hyper}
   \setlength{\tabcolsep}{4pt}
   \centering
 \begin{tabular}{l|c}
 \toprule
  epoch & 100\\
  beginning learning rate & 0.0001\\
  $L_1$ & 8\\
  $L_2$ & 4\\
  spatial vector dimension & 128\\
  day embedding dimension & 16\\
  hour embedding dimension & 16\\
  duration vector dimension & 16\\
  $M$ & 40\\
  $N$ & 5\\
  $\tau$ & 0.8\\
  $\lambda$ & 300\\
  
 \bottomrule
\end{tabular}
\end{table*}

\section{Ablation Study}
To evaluate the effectiveness of the Location Semantics MoE and Personalized MoE modules in NextLocMoE, we conduct ablation studies on Shanghai dataset and further test each component’s transferability in  Shanghai → Kumamoto zero-shot scenario. 
The results are reported in Table~\ref{tab: ablation}.
In the fully-supervised setting, removing the Location Semantics MoE leads to a notable drop in performance, highlighting the crucial role of function-specific experts in capturing complex spatial semantics. 
The impact is even more pronounced when the Personalized MoE is removed—prediction accuracy degrades substantially, underscoring the importance of user persona modeling in enhancing location prediction.
In the zero-shot setting, both ablations result in severe performance degradation.
Notably, the variant without Personalized MoE achieves only 1.437\% Hit@1, indicating that without role-based behavioral modeling, the model becomes significantly less robust to new cities.

\begin{table}[!t]
  \caption{Ablation Study.}
  \label{tab: ablation}
   \setlength{\tabcolsep}{4pt}
 \begin{tabular}{l|ccc|ccc}
    \toprule
    {Method } &  \multicolumn{3}{c|}{Fully-supervised (Shanghai)}   &  \multicolumn{3}{c}{Zero-shot (Shanghai->Kumamoto)}  \\
    & Hit@1 & Hit@5 & Hit@10  & Hit@1 & Hit@5 & Hit@10   \\ 
    \cmidrule(r){1-1}\cmidrule(r){2-4}\cmidrule(r){5-7}
    NextLocMoE & 64.92\% & 75.88\% & 77.43\% & 16.02\% & 36.06\% & 48.42\% \\
    No Location Semantics MoE & 59.72\% & 65.62\% & 69.44\% & 5.519\% & 17.90\% & 28.24\%\\
    No Personalized MoE & 33.68\% & 55.21\% & 63.19\% & 1.437\% & 5.142\% & 9.352\%\\
    \bottomrule
\end{tabular}
\end{table}

%\input{input/check}
%%%%%%%%%%%%%%%%%%%%%%%%%%%%%%%%%%%%%%%%%%%%%%%%%%%%%%%%%%%%

\end{document}